\documentclass[runningheads]{llncs}

\usepackage{accv}

\usepackage{accvabbrv}

\usepackage{graphicx}
\usepackage{booktabs}

\usepackage{multirow}

\usepackage[accsupp]{axessibility}  

\usepackage{hyperref}

\usepackage{orcidlink}

\usepackage[marginal]{footmisc}

\usepackage{arydshln}
\usepackage{bbding}
\usepackage[marginal]{footmisc}

\newcommand{\runinheading}[1]{\par\addvspace\baselineskip{\noindent\bfseries #1}\quad}

\usepackage{arydshln}

\begin{document}

\title{Locate n' Rotate: Two-stage Openable Part Detection with Foundation Model Priors } 



\author{Siqi Li\inst{1,2}  \and
 Xiaoxue Chen \inst{3,4} \and
 Haoyu Cheng \inst{1,2} \and
 Guyue Zhou \inst{3} \and
 Hao Zhao\inst{3} \and
 Guanzhong Tian\inst{1,2(}\Envelope\inst{)}}

\authorrunning{S.~Li et al.}


\institute{Ningbo innovation Center, Zhejiang University.\and
College of Control Science and Engineering, Zhejiang University \and
Institute for AI Industry Research, Tsinghua University \and
Department of Computer Science and Technology Tsinghua University
 }



\maketitle
\footnote{\Envelope\ Corresponding authors

}

\begin{abstract}
Detecting the openable parts of articulated objects is crucial for downstream applications in intelligent robotics, such as pulling a drawer. This task poses a multitasking challenge due to the necessity of understanding object categories and motion. Most existing methods are either category-specific or trained on specific datasets, lacking generalization to unseen environments and objects. In this paper, we propose a Transformer-based Openable Part Detection (OPD) framework named Multi-feature Openable Part Detection (MOPD) that incorporates perceptual grouping and geometric priors, outperforming previous methods in performance. In the first stage of the framework, we introduce a perceptual grouping feature model that provides perceptual grouping feature priors for openable part detection, enhancing detection results through a cross-attention mechanism. In the second stage, a geometric understanding feature model offers geometric feature priors for predicting motion parameters. Compared to existing methods, our proposed approach shows better performance in both detection and motion parameter prediction. Codes and models are publicly available at https://github.com/lisiqi-zju/MOPD
\end{abstract}

\section{Introduction}
\label{sec:intro}

For articulated objects, "openable" refers to an affordance attribute that indicates the parts of objects capable of being opened. For example, a door can be opened through revolute motion, while a drawer can be opened via prismatic motion. Detecting the openable parts within real-world objects is a crucial task in computer vision, with numerous applications in intelligent robotics and manipulation \cite{katz2008manipulating,eisner2022flowbot3d,mo2021where2act,wang2022adaafford}. In this paper, we aim to address the task of Openable Part Detection (OPD), where the input is a single-view image, and the outputs include detected openable parts along with their corresponding motion parameters.

Identifying openable parts within a multi-object scene is a challenging problem, as it can easily be confused by complex indoor furniture. This implies a comprehension of the articulated object's category and function. In addition, the motion parameters of openable parts typically encompass two components: the motion type (prismatic or revolute) and the motion vector consisting of the motion origin and axis. To analyze the motion parameters of openable parts, it's important to establish a geometric understanding of the articulated object's surface.

\begin{figure}[t]
    \centering
    \includegraphics[width=0.95\textwidth]{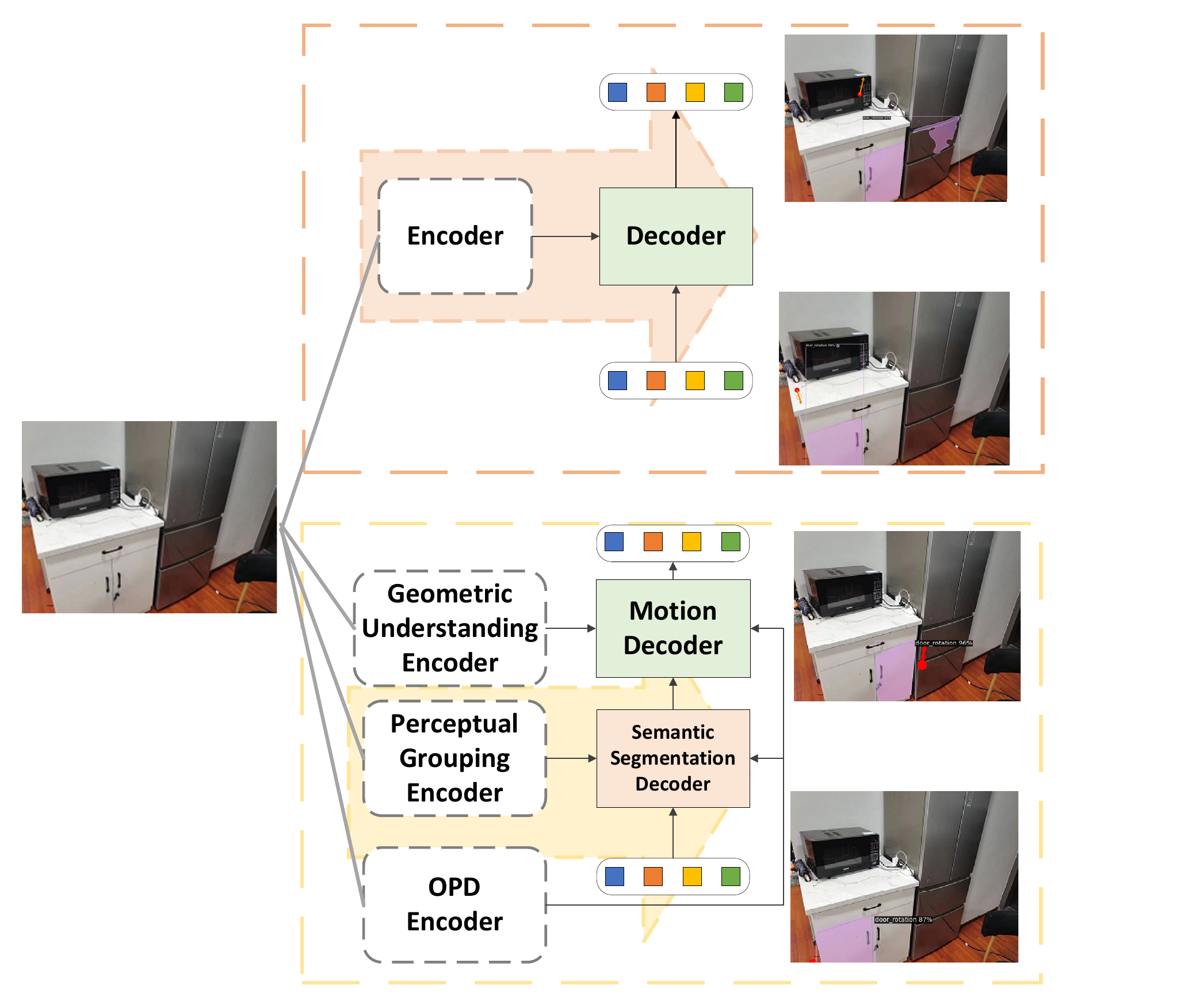}
    \caption{Comparison of network architecture between our framework and MultiOPD. The outputs are the results on an in-the-wild image.  Our model achieves superior performance and showcases generation capabilities to unseen scenarios.}
    \label{fig:teaser}
\end{figure}

With the advent of Embodied AI, a range of methods \cite{hu2018functionality,jiang2022ditto,wang2019shape2motion,yan2020rpm} have emerged to analyze the structure of articulated objects and predict their motion parameters. However, these methods often rely on 3D information inputs such as depth images or point clouds, which offer geometric priors for articulated objects. Moreover, many of these approaches are category-specific, limiting their practicality in the context of intelligent robots. 

Opd  \cite{jiang2022opd}  first introduced a category-agnostic method that predicts openable parts along with their corresponding motion parameters for a single openable object from a single-view image. However, this method only considers scenarios with a single articulated object, which may not be practical in real-world applications. Recently, OPDMulti \cite{sun2023opdmulti} extended OPD to handle multi-object situations. Both methods treat the OPD task as an instance segmentation task and utilize an end-to-end network supervised by ground-truth segmentation masks, without considering the perceptual-level knowledge and geometric priors of the object. This leads to inaccurate predictions of openable parts and their motion parameters. Moreover, the capability of these methods is limited by the training samples and may perform poorly on in-the-wild images, which is illustrated in Fig. \ref{fig:teaser} (the upper one).

To address the aforementioned challenges in openable part detection, we introduce a two-stage Transformer-based framework named MOPD that incorporates the OPD task with perceptual grouping and geometric prior, resulting in superior performance compared to previous methods. Specifically, in the first stage, we introduce a perceptual grouping feature extracted from a perceptual grouping encoder and fuse it with the input image feature using a cross-attention mechanism. This incorporation provides perceptual grouping priors for OPD and enhances detection results. Additionally, in the second stage, we extract a geometric feature from a geometric understanding encoder and fuse it with the first-stage feature, thereby providing geometric priors for motion prediction. It's worth noting that the perceptual grouping and geometric understanding encoder can be replaced with a similar one, making our model a general framework. Fig. \ref{fig:teaser}  provides an architecture comparison with the previous method. And we also introduced a motion cost in the matching step, leveraging our two-stage forecasting approach and combining it with the Optimal Transport model for training. These enhancements played a crucial role in significantly improving the performance of the model.

\section{Related works}
\label{sec:related works}
Our study focuses specifically on the segmentation task of detecting the openable parts of objects, which is named Openable Part Detection (OPD). By integrating instance and geometric features, we attain superior performance compared to prior OPD methods \cite{sun2023opdmulti,jiang2022opd}.Understanding articulated objects is crucial for the development of intelligent robots. Considerable work has been done in this area, including studies such as \cite{ding2024preafford,zhong20233d,chen2022cerberus}. With the rise of Embodied AI, many publicly accessible datasets \cite{abbatematteo2019learning,liu2022akb,martin2019rbo} and physical simulators \cite{deitke2020robothor,shen2021igibson,li2021igibson,savva2019habitat} have been introduced to establish a solid research foundation for articulated objects. To achieve a comprehensive understanding of articulated objects, many works analyze them from various perspectives, such as 3D shape reconstruction \cite{yang2022banmo,yang2021lasr} and 6D pose estimation  \cite{li2020category,wang2019normalized}. With the holistic understanding, some works \cite{katz2008manipulating,eisner2022flowbot3d,mo2021where2act,wang2022adaafford} focus on predict manipulation in a robotic system.  Furthermore, a series of works \cite{jiang2022ditto,wang2019shape2motion,yan2020rpm,sun2023opdmulti,jiang2022opd} have been proposed to analyze parts of articulated objects using RGB images or point clouds, which is crucial for understanding their structures and supporting part-level manipulation. Single-view geometric understanding has received significant attention due to its lower hardware requirements and wider application scope, as evidenced by studies such as \cite{zhao2017physics, zhong2020seeing, long2024adaptive, zheng2024monoocc}.For instance, Ditto \cite{jiang2022ditto}  reconstructs part-level geometry of articulated objects from visual observations of interactions, while OPDMulti \cite{sun2023opdmulti} segments the openable parts of articulated objects from  single-view images using a Transformer-based architecture.

\begin{figure*}[t]
    \centering
    \includegraphics[width=1\textwidth]{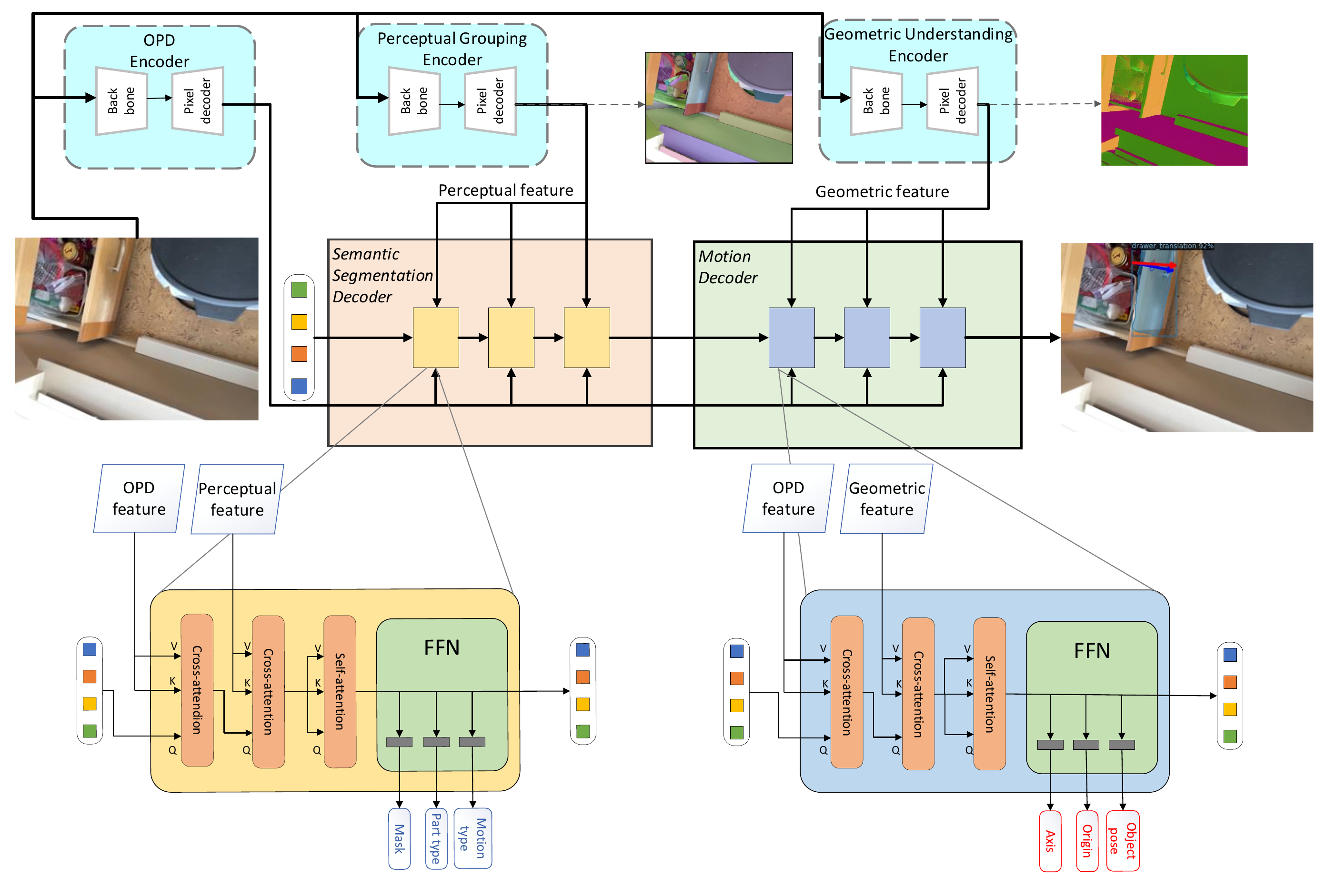}
    \caption{The overall architecture for MOPD. The top side shows the overall network while the bottom shows the decoder in detail. The model employs three encoders to extract the features from the images. The pixel-level embeddings from the encoder are passed to the transformer decoder with learnable part queries to learn embeddings that are used to predict the openable part. The OPD feature and perceptual grouping feature are successively crossed in the segmentation decoder to obtain a high-resolution mask. In the same way, the OPD feature and geometric feature are used in the motion decoder. The motion type, part type, and mask are predicted in all FFN layers of the semantic segmentation decoder, while the object poses, origin, and axis are predicted in all FFN layers of the motion decoder. The image on the far right shows the output. The GT axis is in blue and the predicted axis is in red.}
    \label{fig:framework}
\end{figure*}

\section{Methods}
\label{sec:methods}

In this paper, our aim is to detect the openable parts of articulated objects based on RGB images. In this section, we begin by formally defining the Openable Part Detection (OPD) task in Section~\ref{sec:OPD Task}. Following that, we provide an overview of the overall framework in Section~\ref{sec:Network Architecture}. Additionally, we introduce the details of multi-feature for perceptual grouping and geometric understanding in Section~\ref{sec:Multi Feature}. We then delve into the network architectures in Section~\ref{sec:Transformer Decoder} and~\ref{sec:Losses and FFN layer}. Finally, we introduce the our matching strategy with optimal transport assignment.

\subsection{Preliminary for OPD Task}\label{sec:OPD Task}

The objective of openable part detection is to identify all openable components from  single vision. Unlike traditional instance segmentation tasks, which focus on delineating individual objects, openable part detection requires a deeper understanding of the concept of "openable". While this concept is intuitive for humans, it is challenging to precisely define. Therefore, we simplify the notion of openable by decomposing it into distinct constituent parts within the dataset following \cite{sun2023opdmulti}. Specifically, an openable part ${p}_{i}$ is characterized by a 2D bounding box ${b}_{i}$ or a segmentation mask ${m}_{i}$, along with a motion axis direction ${a}_{i} \in R^{3}$ and motion origin ${o}_{i} \in R^{3}$ (for revolute motion type only). Each openable part is categorized into one of three semantic types ${l}_{i} \in \{drawer, door, lid\} $and one of two motion types ${c}_{i} \in  \{prismatic, revolute\}$. Since there is usually more than one articulated object in a real-world indoor scene, an output from an image will consist of a set of openable parts ${P = \{{p}_{i},...,{p}_{k}\}}$.

\subsection{Overall Network Architecture}\label{sec:Network Architecture}

To address the Openable Part Detection (OPD) task, we introduce a novel methodology called MOPD (Multi-feature Openable Part Detection), which is based on Mask2former \cite{cheng2022masked}. In the upcoming paragraphs, we will delve into the differences between our approach and Mask2former to highlight the contributions of our OPD model.

The network structure of Mask2former mainly includes three stages: multi-level feature extraction based on Backbone, a pixel-level decoder based on a multi-level deformable self-attention mechanism, and a multi-head cross-attention mechanism decoder based on the mask. In MOPD, we propose two key network architecture contributions to enhance the performance of the OPD task:

\begin{enumerate}
\item In OPD task the motion parameters are predicted with detection together, by the same query through the different model heads. It made the two kinds of prediction disturb each other. So we predict them in two decoders which incorporate different features. To incorporate perceptual grouping and geometric priors, two types of additional features are introduced through specialized feature encoders. These feature encoders comprise a backbone and a pixel-level decoder. The input image is processed to obtain these features, which subsequently serve as the key and value vectors for input into the transformer decoder. The details will be depicted in the next section.

\item In order to address the OPD task, we propose a two-stage task-specific framework, each corresponding to perceptual grouping and geometric features respectively. In the first stage, we emphasize the semantic information of the objects and predict bounding boxes (or masks), part types, and motion types. In the second stage, we incorporate geometric features and output object poses, origins, and axes. Through different FNN layers, the two categories of predictions are generated in different modules successively.
\end{enumerate}

The overall model architecture is depicted in Fig.~\ref{fig:framework}. The model to maintain real-time utilizes a ResNet-50 backbone as the Openable Part Detection (OPD) encoder to extract features from the input image. Additionally, it employs a perceptual grouping feature encoder sourced from  EfficientSAM \cite{xiong2023efficientsam}, along with a geometric understanding encoder from  DSINE \cite{bae2024dsine}. Both the perceptual grouping encoder and geometric understanding encoder are pre-trained on their respective tasks and then fine-tuned with the entire model for the openable part detection task. The features extracted from the perceptual grouping encoder and geometric understanding encoder are fused with the image features using a cross-attention layer within the transformer.

\subsection{Multi-Feature of Perceptual grouping and Geometric}\label{sec:Multi Feature}
Given the perceptual grouping and geometric nature of the OPD task, our approach focuses on enhancing model performance and generalization by incorporating perceptual grouping and geometric features learned from other computer vision tasks. These features are extracted from models pre-trained in their respective tasks, and we fine-tune their encoders specifically for the OPD task.
Specifically, we introduce two types of encoders to aid in detecting openable parts: a perceptual grouping encoder to help the model understand the categories of articulated objects and a geometric understanding encoder to produce spatial features to assist in predicting the origin, axis, and object pose.
These encoders are combined with the backbone and pixel decoder. The image passes through the backbone to obtain the embedding, and then through the pixel decoder to extract the feature. These features are utilized as a key and value vector in the cross-attention layer to fuse with the query vector obtained by the OPD encoder. Perceptual grouping encoders and  geometric encoders  are respectively be pretrained by EfficientSAM and DSINE.

\textbf{EfficientSAM}: A lightweight SAM model that exhibits decent performance with largely reduced complexity. It takes SAM pre-trained lightweight image encoders and mask decoder to build EfficientSAMs and finetune the models on SA-1B for segment anything task\cite{xiong2023efficientsam}.

\textbf{DSINE}: It utilizes the per-pixel ray direction and encodes the relationship between neighboring surface normals by learning their relative rotation. It shows a stronger generalization ability, despite being trained on an orders of magnitude smaller dataset \cite{bae2024dsine}.

\begin{figure*}[ht]
\centering
\resizebox{0.8\linewidth}{!}{%
\begin{tabular}{ccccccc}
\toprule
OPDMulti & \begin{minipage}[b]{0.3\columnwidth}\centering\raisebox{-.5\height}{\includegraphics[width=\linewidth]{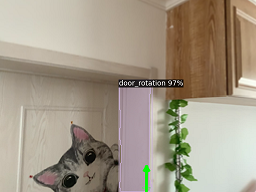}}\end{minipage}
         & Miss
         & Miss
         & Miss
         & Miss\\
MOPD     & \begin{minipage}[b]{0.3\columnwidth}\centering\raisebox{-.5\height}{\includegraphics[width=\linewidth]{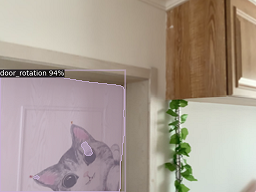}}\end{minipage}
         & \begin{minipage}[b]{0.3\columnwidth}\centering\raisebox{-.5\height}{\includegraphics[width=\linewidth]{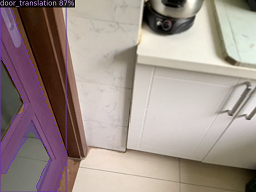}}\end{minipage}
         & \begin{minipage}[b]{0.3\columnwidth}\centering\raisebox{-.5\height}{\includegraphics[width=\linewidth]{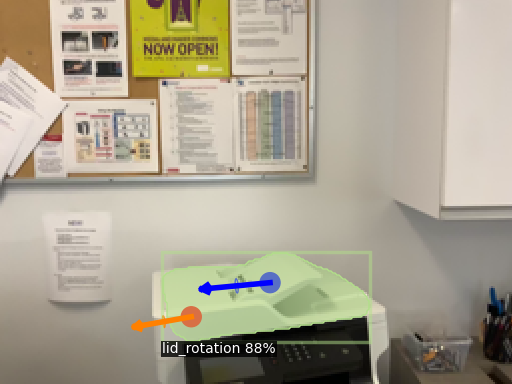}}\end{minipage} 
         & \begin{minipage}[b]{0.3\columnwidth}\centering\raisebox{-.5\height}{\includegraphics[width=\linewidth]{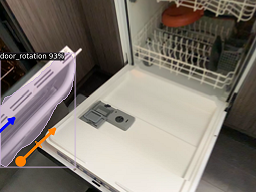}}\end{minipage}
         & \begin{minipage}[b]{0.3\columnwidth}\centering\raisebox{-.5\height}{\includegraphics[width=\linewidth]{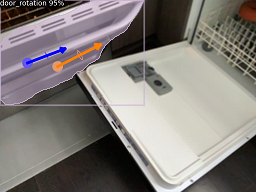}}\end{minipage}\\
\midrule
OPDMulti & \begin{minipage}[b]{0.3\columnwidth}\centering\raisebox{-.5\height}{\includegraphics[width=\linewidth]{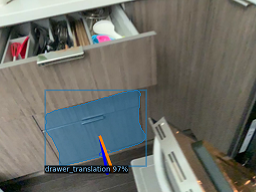}}\end{minipage}
         & \begin{minipage}[b]{0.3\columnwidth}\centering\raisebox{-.5\height}{\includegraphics[width=\linewidth]{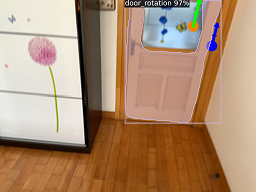}}\end{minipage}
         & \begin{minipage}[b]{0.3\columnwidth}\centering\raisebox{-.5\height}{\includegraphics[width=\linewidth]{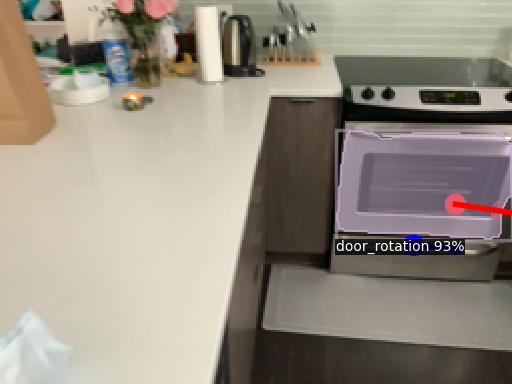}}\end{minipage} 
         & \begin{minipage}[b]{0.3\columnwidth}\centering\raisebox{-.5\height}{\includegraphics[width=\linewidth]{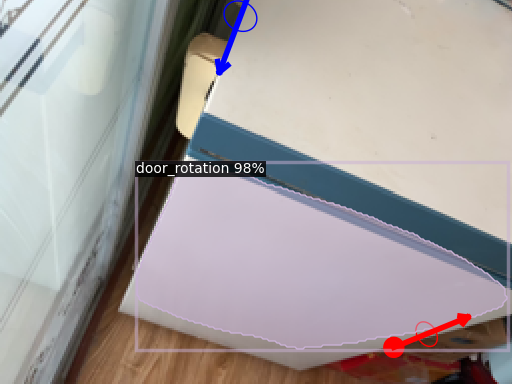}}\end{minipage}
         & \begin{minipage}[b]{0.3\columnwidth}\centering\raisebox{-.5\height}{\includegraphics[width=\linewidth]{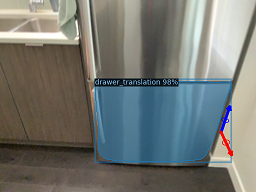}}\end{minipage}\\
MOPD     & \begin{minipage}[b]{0.3\columnwidth}\centering\raisebox{-.5\height}{\includegraphics[width=\linewidth]{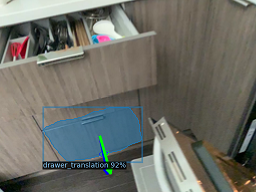}}\end{minipage}
         & \begin{minipage}[b]{0.3\columnwidth}\centering\raisebox{-.5\height}{\includegraphics[width=\linewidth]{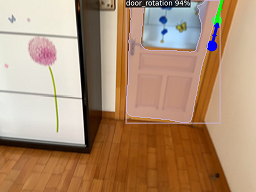}}\end{minipage}
         & \begin{minipage}[b]{0.3\columnwidth}\centering\raisebox{-.5\height}{\includegraphics[width=\linewidth]{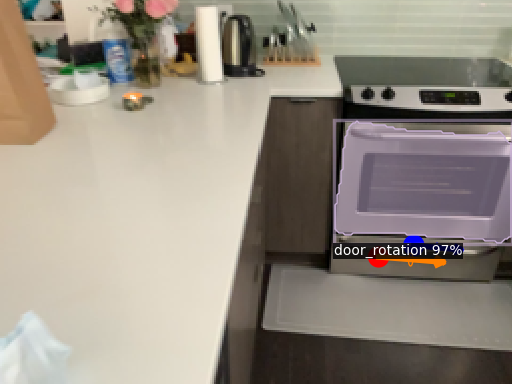}}\end{minipage} 
         & \begin{minipage}[b]{0.3\columnwidth}\centering\raisebox{-.5\height}{\includegraphics[width=\linewidth]{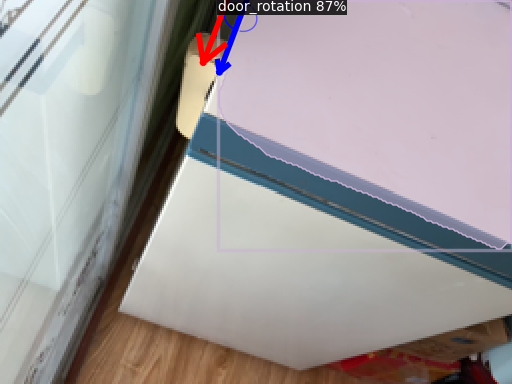}}\end{minipage}
         & \begin{minipage}[b]{0.3\columnwidth}\centering\raisebox{-.5\height}{\includegraphics[width=\linewidth]{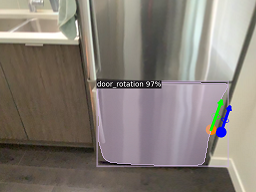}}\end{minipage}\\
\midrule
OPDMulti & \begin{minipage}[b]{0.3\columnwidth}\centering\raisebox{-.5\height}{\includegraphics[width=\linewidth]{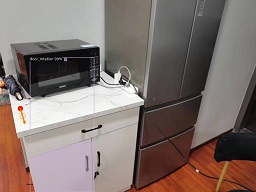}}\end{minipage}
         & \begin{minipage}[b]{0.3\columnwidth}\centering\raisebox{-.5\height}{\includegraphics[width=\linewidth]{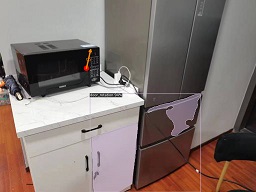}}\end{minipage}
         & \begin{minipage}[b]{0.3\columnwidth}\centering\raisebox{-.5\height}{\includegraphics[width=\linewidth]{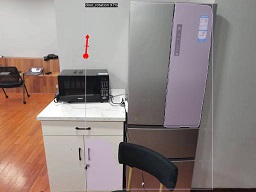}}\end{minipage} 
         & \begin{minipage}[b]{0.3\columnwidth}\centering\raisebox{-.5\height}{\includegraphics[width=\linewidth]{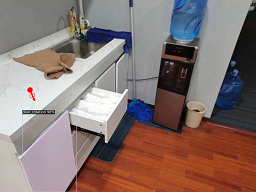}}\end{minipage}
         & Miss\\
MOPD     & \begin{minipage}[b]{0.3\columnwidth}\centering\raisebox{-.5\height}{\includegraphics[width=\linewidth]{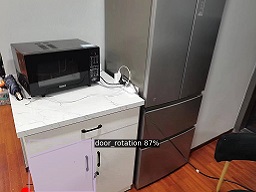}}\end{minipage}
         & \begin{minipage}[b]{0.3\columnwidth}\centering\raisebox{-.5\height}{\includegraphics[width=\linewidth]{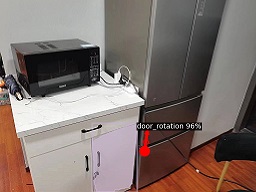}}\end{minipage}
         & \begin{minipage}[b]{0.3\columnwidth}\centering\raisebox{-.5\height}{\includegraphics[width=\linewidth]{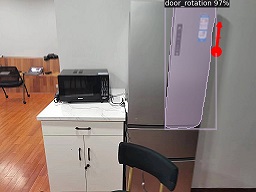}}\end{minipage} 
         & \begin{minipage}[b]{0.3\columnwidth}\centering\raisebox{-.5\height}{\includegraphics[width=\linewidth]{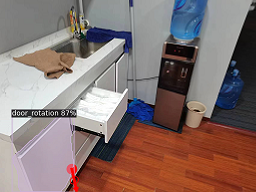}}\end{minipage}
         & \begin{minipage}[b]{0.3\columnwidth}\centering\raisebox{-.5\height}{\includegraphics[width=\linewidth]{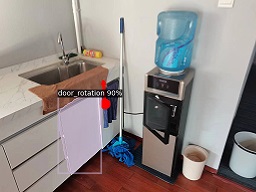}}\end{minipage}\\
\bottomrule
\end{tabular}
}
\caption{Qualitative results on the OPDMulti and MOPD val split. The first two rows are a comparison of MOPD variants with OPDMulti in valid dataset. The last rows are a  comparison in the wild. The GT axis is in blue and the predicted axis is in green if it is within ${5^{\circ}}$ of the GT,orange if between ${5^{\circ}}$and ${10^{\circ}}$ and red if the angle difference is greater than ${10^{\circ}}$.}\label{fig:result}
\end{figure*}

\subsection{Transformer Decoder}\label{sec:Transformer Decoder}
To obtain high-resolution masks, we employ a two-stage strategy utilizing multi-scale visual feature maps at increasing resolutions, each of which is respectively inputted into the perceptual grouping and geometric understanding decoder.  The transformer comprises two decoders: one for semantic segmentation stacked for $L_1$ layers, and the other for motion prediction stacked for $L_2$ layers. Each layer in the transformer stack consists of two cross-attention layers, one self-attention layer, and a feedforward neural network (FFN) layer. The mask and type prediction are separated from the motion prediction. By employing different combinations of prediction heads, the prediction position can be flexibly adjusted at any position in the decoder layers. A comprehensive overview of the architecture is depicted in Fig. \ref{fig:framework}.

\subsection{FFN layer and Training Losses}\label{sec:Losses and FFN layer}
For the segmentation and motion losses, we add the auxiliary loss after each transformer decoder. And we predict them successively in two different transformer decoders with different features.
\begin{enumerate}
\item  \textbf{Segmentation losses}: The mask segmentation loss for the first stage comprises the following components: binary cross-entropy loss ${L}_{ce}$, the dice loss ${L}_{dice}$, cross-entropy loss ${L}_{cls}$ and the motion type cross-entropy loss ${L}_{c}$. The overall formulation is: ${L}_{seg}={\lambda}_{cd} {L}_{ce} + {\lambda}_{dice} {L}_{dice} + {\lambda}_{cls} + {L}_{cls} + {\lambda}_{c} {L}_{c}$, where ${\lambda}_{\cdot}$ is the loss weight.

\item  \textbf{Motion losses}: The motion losses for the second stage consist of the following components: smooth L1 losses for the motion axis ${L}_{a}$,  motion origin losses ${L}_{o}$ and object pose losses ${L}_{o}$. ${L}_{mot}={\lambda}_{a} {L}_{a} + {\lambda}_{o} {L}_{o} + {\lambda}_{pose} {L}_{pose}$, where $\lambda_{a}$, $\lambda_{o}$, and $\lambda_{pose}$ are the weighting coefficients for each respective loss component.

\end{enumerate}
We sum the segmentation loss and the motion loss to obtain the overall loss used during training: $L={L}_{seg} + {L}_{mot}$.

\subsection{Optimal Transport}\label{sec:Optimal Transport}
Traditional object detectors perform detection by predicting classification labels and regression offsets for a set of proposals. To train the detector, matching targets for each proposal is a necessary process. Most strategies may result in sub-optimal proposal assignments for each ground truth individually without context, as assigning ambiguous proposals to any ground truth might bring harmful gradients to other ground truths. To achieve a globally optimal assignment result in a one-to-many situation, optimal transport formulates label assignment as an Optimal Transport (OT) problem. Specifically, The cost between the ground truth and a proposal is defined solely by their pairwise classification cost. After formalizing this, finding the optimal assignment scheme is transformed into solving the optimal transport plan, which can be efficiently and quickly solved using the ready-made Sinkhorn-Knopp iteration. We name this assignment strategy Optimal Transport Assignment (OTA).\par
Previous works keep working on the object matching, lacking the attention of matching motion itself. To address the influence of motion matching, we propose the match cost of motion. Our proposed match cost includes the following two aspects:
\begin{enumerate}
\item  \textbf{Origin Match cost}: We set the origin match cost as the normalized cross product of the predicted origin drift and the ground truth axis:
\begin{center}
    ${C}_{origin} = \frac{(\mathbf{O}_{pred} - \mathbf{O}_{gt}) \times \mathbf{I}_{gt}}{{L}_{diag}}$
\end{center}
where ${L}_{diag}$ is the diagonal length of the object, characterizing the size.

\item  \textbf{Axis Match cost}: The axis match cost is the angular difference between the predicted axis and the ground truth axis:
\begin{center}
    ${C}_{axis} = arccos(\frac{{I}_{pred} \cdot {I}_{gt}}{|{I}_{pred}| |{I}_{gt}|})$
\end{center}

\end{enumerate}
We sum the origin match cost and the axis match cost to obtain the overall matching cost matrix used during training: $C={C}_{obj} + {C}_{origin} + {C}_{axis}$.${C}_{obj}$ represents the matching cost matrix in a traditional detection task.

\section{Experiments}
\label{sec:experiments}
In this section, we conduct experiments to verify the effectiveness of our model and compare MOPD and several varieties with the previous baselines. We also show the efficiency of our algorithm with different modules through ablation studies.

\subsection{Evaluation Metrics}
We follow the evaluation metrics for part detection and motion prediction used in OPDmulti \cite{sun2023opdmulti}. The metrics extend the traditional mAP metric. To evaluate the detection of openable parts the metrics include several metrics. First is AP@IoU=0.5 for the predicted part label and 2D bounding box (PDet) or mask. On the basis of PDet to evaluate the motion parameters, For each metric, the detection is further constrained by whether: the motion type is matched (+M), motion type and motion axis are matched (+MA), and whether the motion type, axis, and origin are all matched (+MAO), within predefined error thresholds \cite{sun2023opdmulti}.

\begin{table}[]
\centering
\caption{Quantitative results. The MOPD model utilizes EfficienceSAM and DSINE as the perceptual grouping encoder and geometric understanding encoder. }
\resizebox{0.5\linewidth}{!}{
\begin{tabular}{c|cccc}
\toprule 
\multirow{2}{*}{\textbf{Model}} &  \multicolumn{4}{c}{Part-averaged mAP  $\% \uparrow$}       \\ & PDet & +M & +MA & +MAO \\
\midrule
OPDRCNN-C               & 27.3               & 25.7            & 8.8                & 7.8             \\
OPDRCNN-O               & 20.0               & 18.3            & 3.9                & 0.5             \\
OPDRCNN-P               & 20.9               & 19.0            & 7.2                & 5.7             \\
OPDFORMER-C             & 30.3               & 28.9            & 13.1               & 12.1            \\
OPDFORMER-O             & 30.1               & 28.5            & 5.2                & 1.6             \\
OPDFORMER-P(OPDMulti)   & 32.9               & 31.6            & 19.4               & 16.0            \\
\hdashline
MOPD                    & 37.3               & 36.1            & \textbf{20.7}      & 16.6            \\
MOPD(Optimal Transport) & \textbf{37.8}      & \textbf{37.7}   & 20.1               & \textbf{17.2}   \\
\bottomrule
\end{tabular}
}
\label{table:Comparison of result}
\end{table}

\subsection{Implementation details}

Our model is implemented based on Mask2Former \cite{cheng2022masked}. We employ a ResNet-50 backbone pre-trained on the COCO dataset  \cite{lin2014microsoft}, with a learning rate of 0.0001. Experiments are conducted on the OPDMulti dataset \cite{sun2023opdmulti} using an A100-SXM4-80GB GPU. Models evaluated on the OPDMulti dataset are initially trained on the OPDReal dataset \cite{jiang2022opd}, followed by fine-tuning on the OPDMulti dataset. Training is performed end-to-end for 60000 steps, and the best checkpoint is selected based on validation set performance (using the +MAO metric). A confidence threshold of 0.8 is applied to determine the validity of predicted parts.

\begin{figure*}[ht]
\centering
\resizebox{\linewidth}{!}{%
\begin{tabular}{ccccccc}
\toprule
MOPD(w/o perceptual) & \begin{minipage}[b]{0.3\columnwidth}\centering\raisebox{-.5\height}{\includegraphics[width=\linewidth]{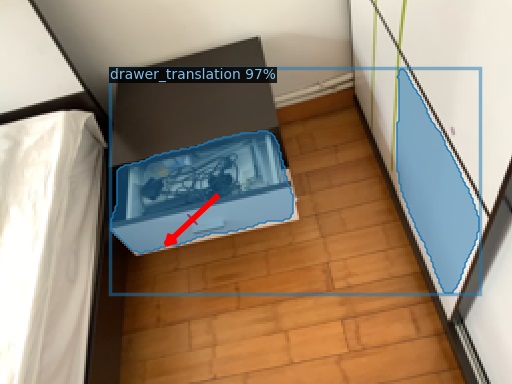}}\end{minipage}
         & \begin{minipage}[b]{0.3\columnwidth}\centering\raisebox{-.5\height}{\includegraphics[width=\linewidth]{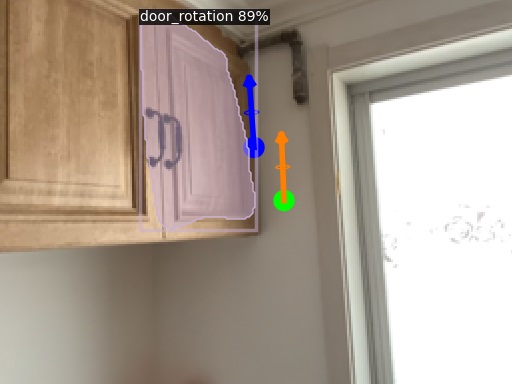}}\end{minipage}
         & \begin{minipage}[b]{0.3\columnwidth}\centering\raisebox{-.5\height}{\includegraphics[width=\linewidth]{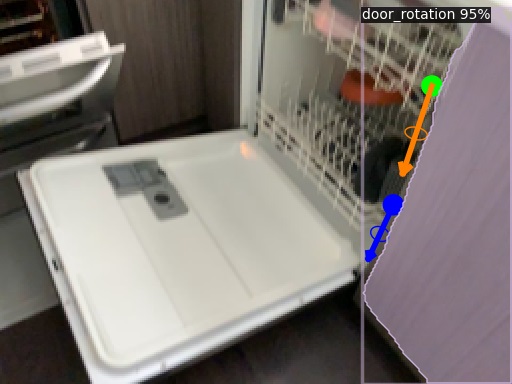}}\end{minipage} 
         & Miss
         & Miss
         \\
MOPD   & \begin{minipage}[b]{0.3\columnwidth}\centering\raisebox{-.5\height}{\includegraphics[width=\linewidth]{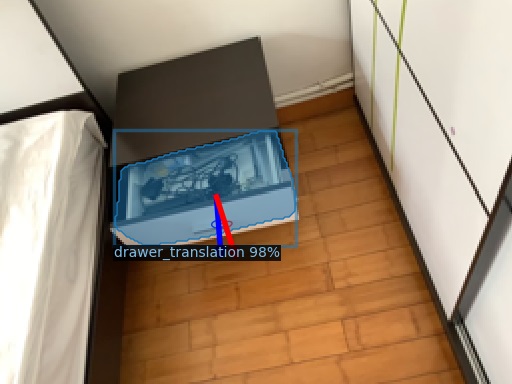}}\end{minipage}
         & \begin{minipage}[b]{0.3\columnwidth}\centering\raisebox{-.5\height}{\includegraphics[width=\linewidth]{figures_ab/34-4080__1.jpg}}\end{minipage}
         & \begin{minipage}[b]{0.3\columnwidth}\centering\raisebox{-.5\height}{\includegraphics[width=\linewidth]{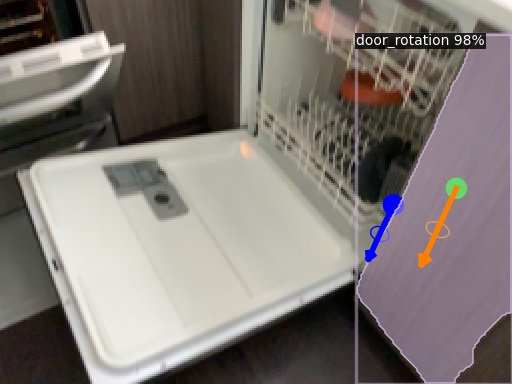}}\end{minipage} 
         & \begin{minipage}[b]{0.3\columnwidth}\centering\raisebox{-.5\height}{\includegraphics[width=\linewidth]{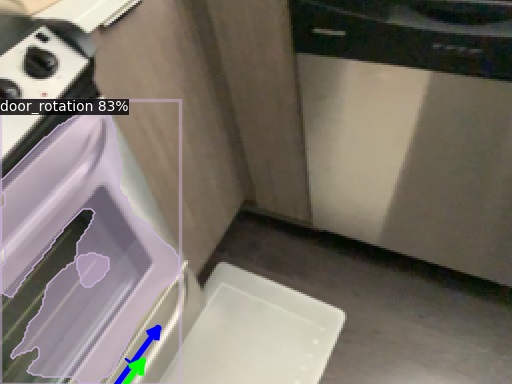}}\end{minipage}
         & \begin{minipage}[b]{0.3\columnwidth}\centering\raisebox{-.5\height}{\includegraphics[width=\linewidth]{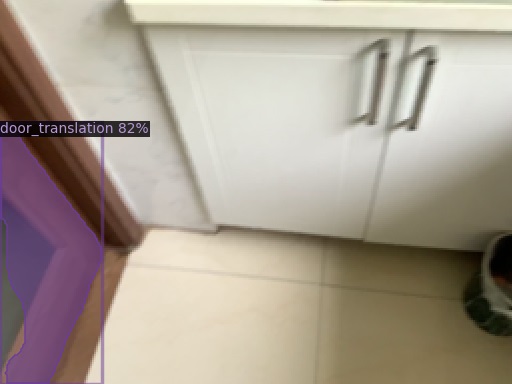}}\end{minipage}\\
\midrule
EfficientSAM    & \begin{minipage}[b]{0.3\columnwidth}\centering\raisebox{-.5\height}{\includegraphics[width=\linewidth]{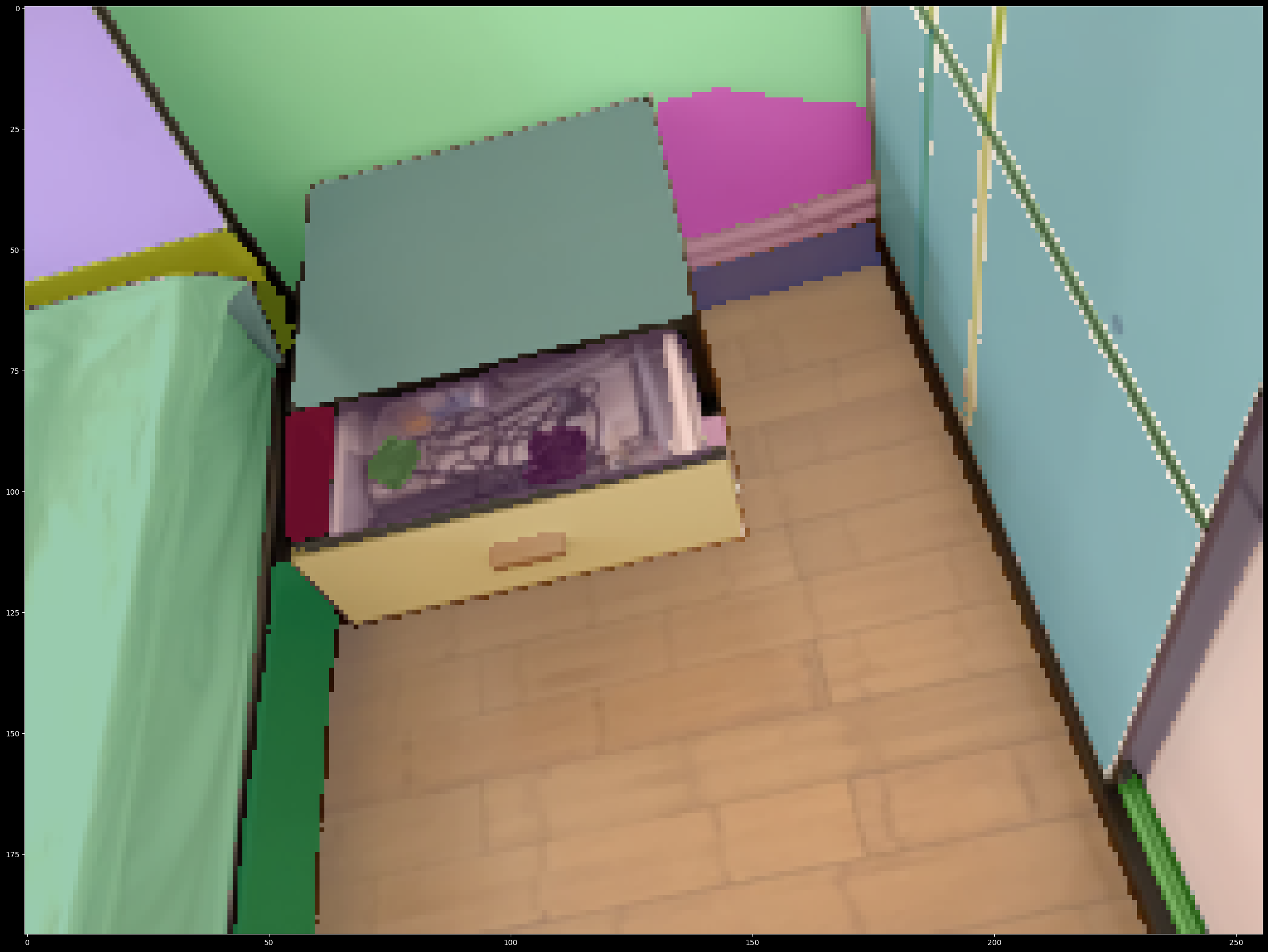}}\end{minipage}
         & \begin{minipage}[b]{0.3\columnwidth}\centering\raisebox{-.5\height}{\includegraphics[width=\linewidth]{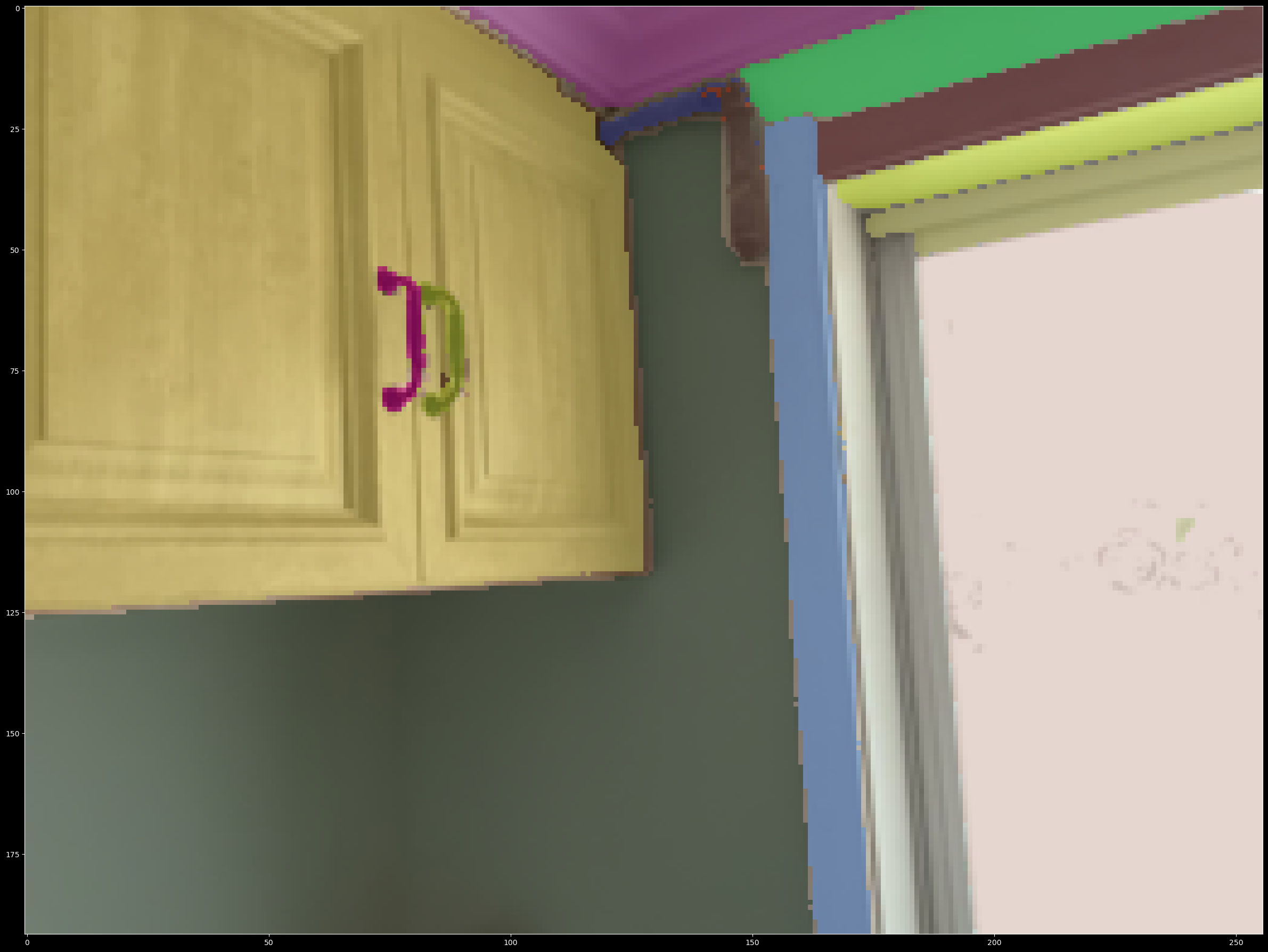}}\end{minipage}
         & \begin{minipage}[b]{0.3\columnwidth}\centering\raisebox{-.5\height}{\includegraphics[width=\linewidth]{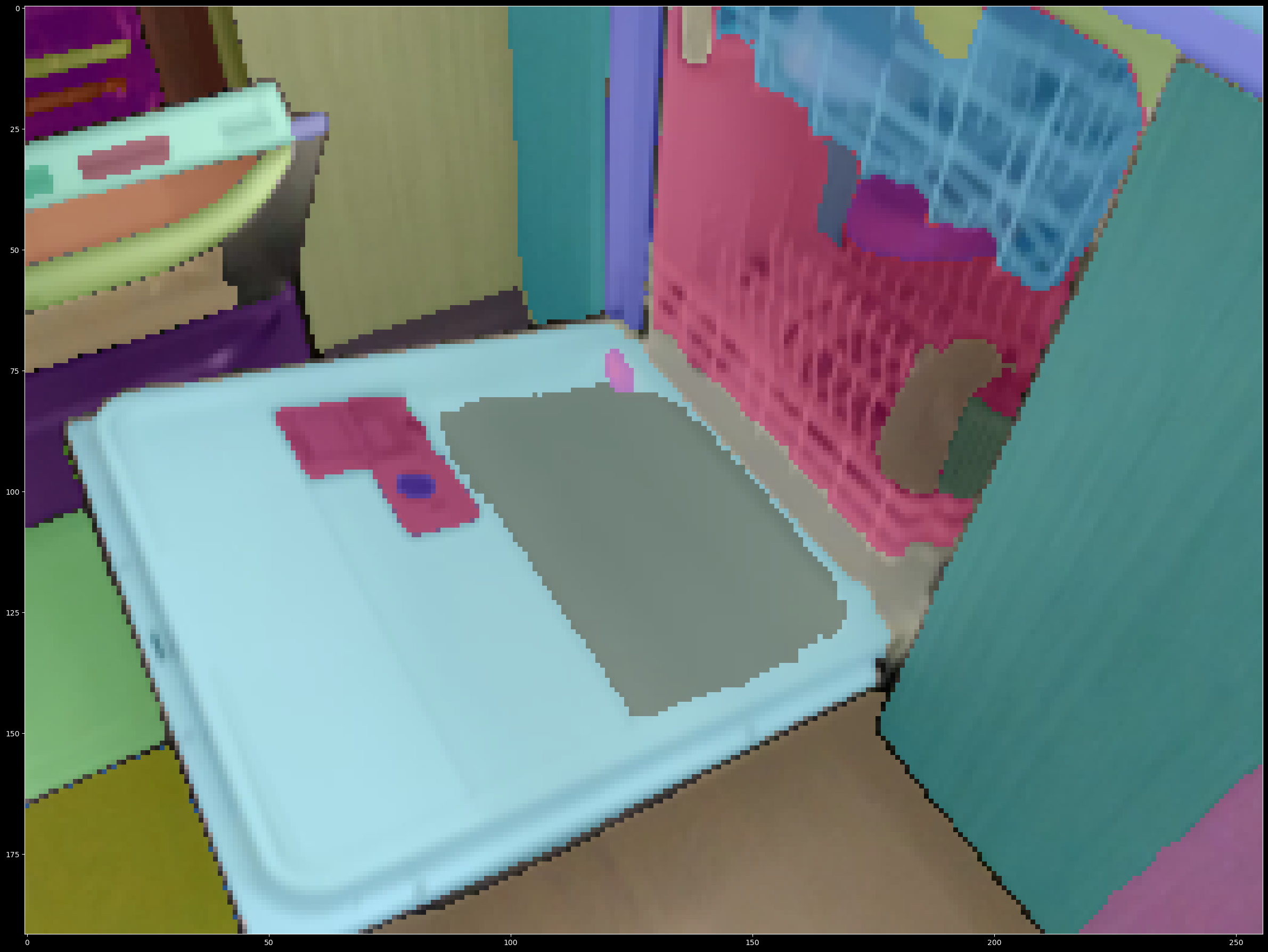}}\end{minipage} 
         & \begin{minipage}[b]{0.3\columnwidth}\centering\raisebox{-.5\height}{\includegraphics[width=\linewidth]{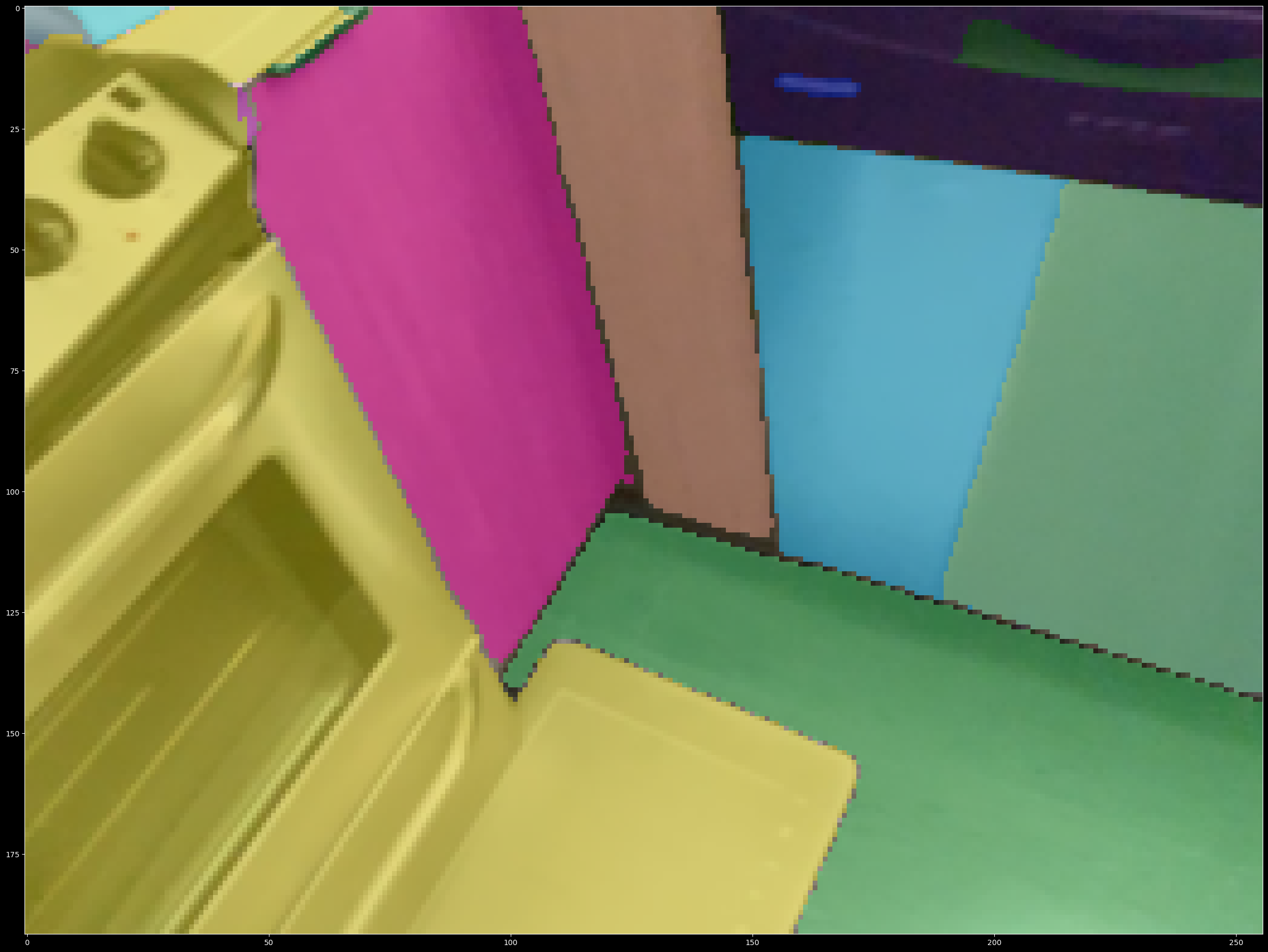}}\end{minipage}
         & \begin{minipage}[b]{0.3\columnwidth}\centering\raisebox{-.5\height}{\includegraphics[width=\linewidth]{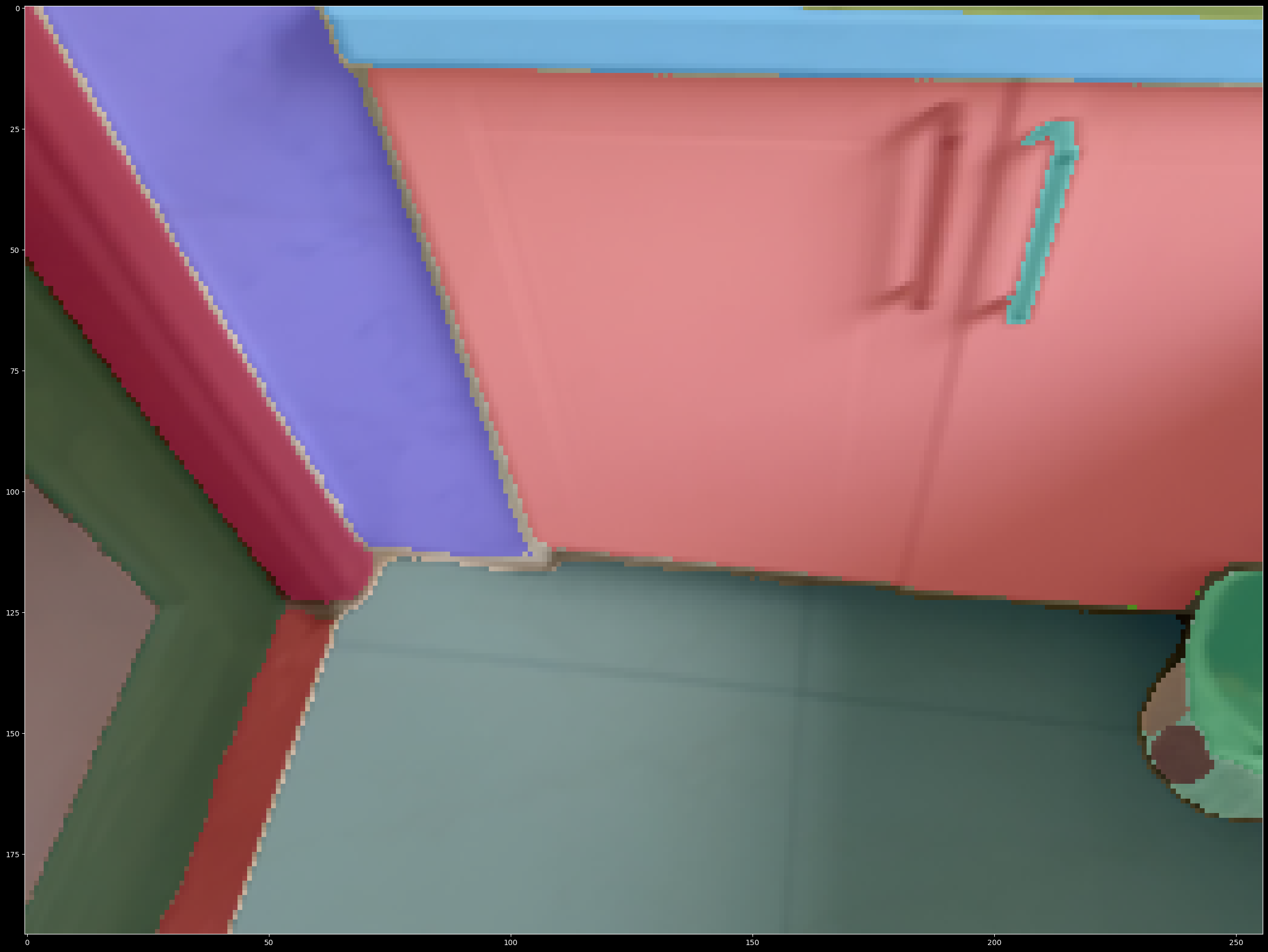}}\end{minipage}\\
\bottomrule
\end{tabular}
}
\caption{At the top, there is a comparison of the results obtained from the w/o  perceptual grouping encoder in MOPD. At the bottom, there is the output from EfficientSAM, which we utilized to pre-train the encoder. The figure demonstrate that our model indeed utilizes the pre-trained encoder. Since the DETR is a query-based model, it can occasionally detect two distinct objects as a single entity.   However, by leveraging the segmentation capabilities inherent in the EfficientSAM model, we are able to effectively mitigate such errors and improve the overall accuracy of detections. The quantitative result are shown in Table \ref{table:efficiency1}.
}\label{fig:ablation study SAM}
\end{figure*}

\begin{figure*}[ht]
\centering
\resizebox{\linewidth}{!}{%
\begin{tabular}{ccccccc}
\toprule
MOPD(w/o geometric) & \begin{minipage}[b]{0.3\columnwidth}\centering\raisebox{-.5\height}{\includegraphics[width=\linewidth]{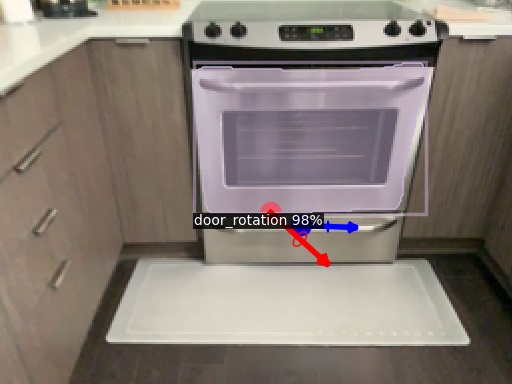}}\end{minipage}
         & \begin{minipage}[b]{0.3\columnwidth}\centering\raisebox{-.5\height}{\includegraphics[width=\linewidth]{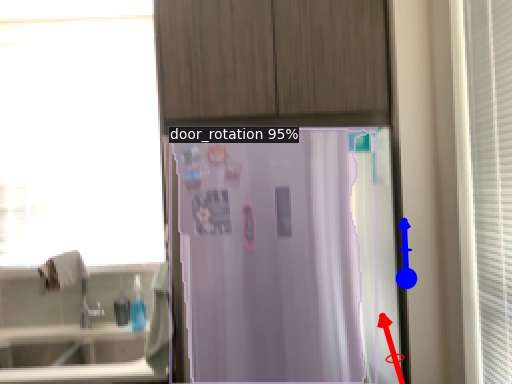}}\end{minipage}
         & \begin{minipage}[b]{0.3\columnwidth}\centering\raisebox{-.5\height}{\includegraphics[width=\linewidth]{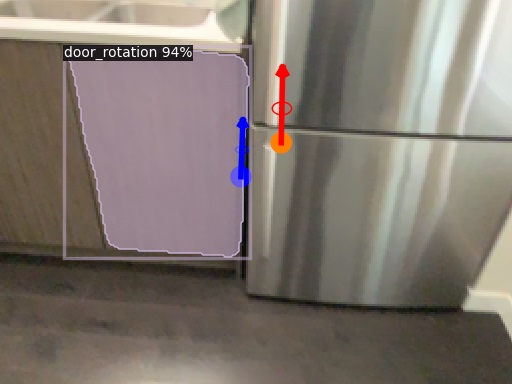}}\end{minipage} 
         & \begin{minipage}[b]{0.3\columnwidth}\centering\raisebox{-.5\height}{\includegraphics[width=\linewidth]{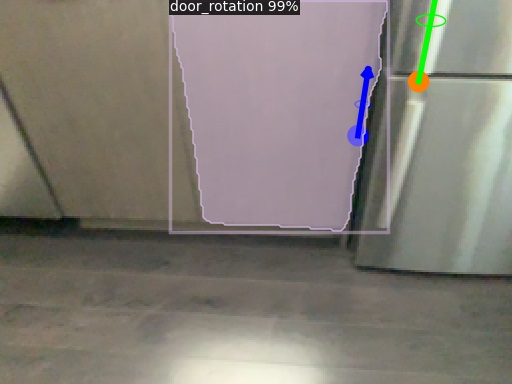}}\end{minipage}
         & \begin{minipage}[b]{0.3\columnwidth}\centering\raisebox{-.5\height}{\includegraphics[width=\linewidth]{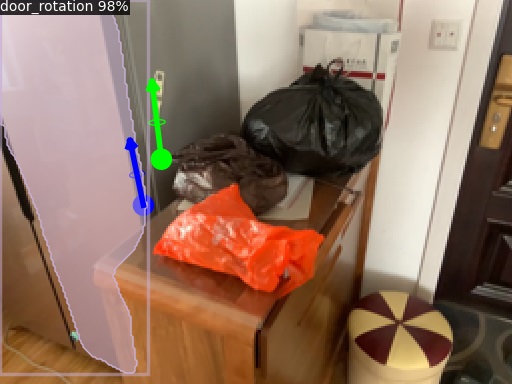}}\end{minipage}\\
MOPD    & \begin{minipage}[b]{0.3\columnwidth}\centering\raisebox{-.5\height}{\includegraphics[width=\linewidth]{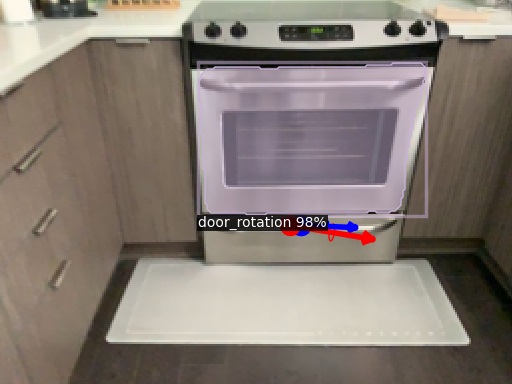}}\end{minipage}
         & \begin{minipage}[b]{0.3\columnwidth}\centering\raisebox{-.5\height}{\includegraphics[width=\linewidth]{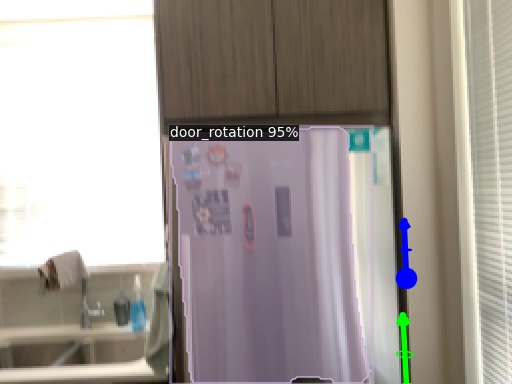}}\end{minipage}
         & \begin{minipage}[b]{0.3\columnwidth}\centering\raisebox{-.5\height}{\includegraphics[width=\linewidth]{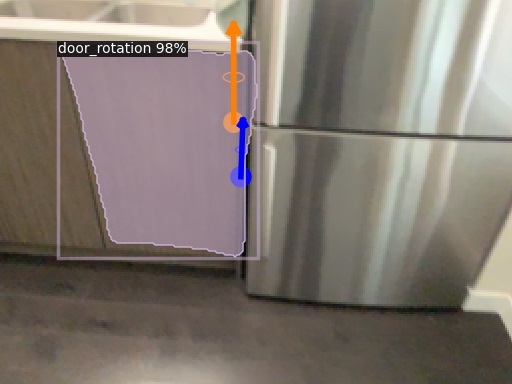}}\end{minipage} 
         & \begin{minipage}[b]{0.3\columnwidth}\centering\raisebox{-.5\height}{\includegraphics[width=\linewidth]{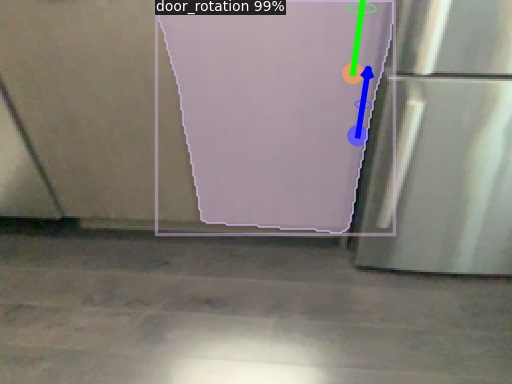}}\end{minipage}
         & \begin{minipage}[b]{0.3\columnwidth}\centering\raisebox{-.5\height}{\includegraphics[width=\linewidth]{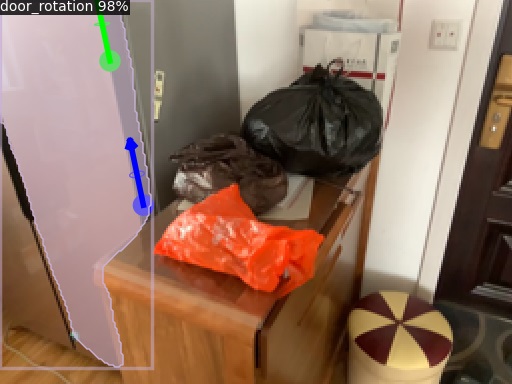}}\end{minipage}\\
\midrule
DSINE   & \begin{minipage}[b]{0.3\columnwidth}\centering\raisebox{-.5\height}{\includegraphics[width=\linewidth]{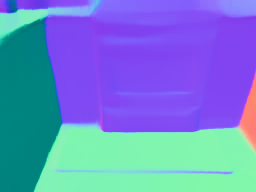}}\end{minipage}
         & \begin{minipage}[b]{0.3\columnwidth}\centering\raisebox{-.5\height}{\includegraphics[width=\linewidth]{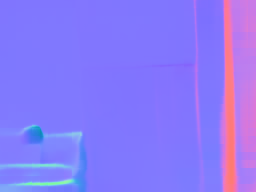}}\end{minipage}
         & \begin{minipage}[b]{0.3\columnwidth}\centering\raisebox{-.5\height}{\includegraphics[width=\linewidth]{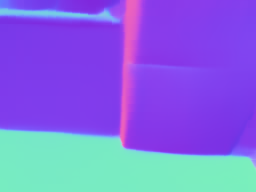}}\end{minipage} 
         & \begin{minipage}[b]{0.3\columnwidth}\centering\raisebox{-.5\height}{\includegraphics[width=\linewidth]{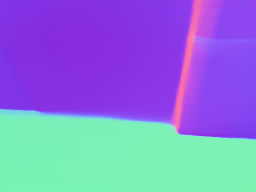}}\end{minipage}
         & \begin{minipage}[b]{0.3\columnwidth}\centering\raisebox{-.5\height}{\includegraphics[width=\linewidth]{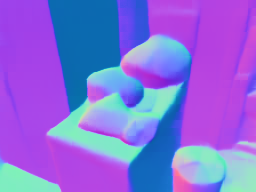}}\end{minipage}\\
\bottomrule 
\end{tabular}
}
\caption{At the top, there is a comparison of the results obtained from the w/o  geometric encoder in MOPD. At the bottom, there is the output from  DSINE, which we utilized to pre-train the geometric encoder. Through the plugging in of geometric features, the model corrects the axis direction to make it closer to the surface normal evaluation. It indicates that the decoder indeed takes advantage of geometric features. When comparing the two models, we can observe that our model has better precision with origin prediction, especially when the two models have a similar axis prediction.  This is because the RGB picture lacks information regarding the degree of the two crossing surfaces, which makes the model unable to provide an accurate prediction of Three-dimensional coordinates when the axis is near the edge of the door and lid.  The introduction of normal features can alleviate this issue by pushing the origin away from the incorrect surface.}\label{fig:ablation study normal}
\end{figure*}

\begin{figure*}
  \centering
  \subfloat{
    \includegraphics[width=0.4\linewidth]{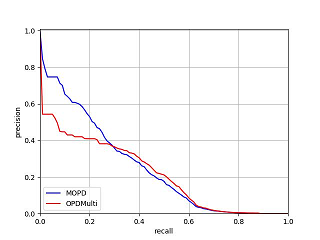}
  }
  \subfloat
  {
    \includegraphics[width=0.4\linewidth]{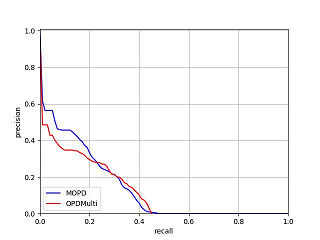}
  }

  \subfloat{
    \includegraphics[width=0.4\linewidth]{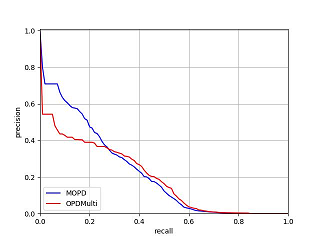}
  }
  \subfloat
  {
    \includegraphics[width=0.4\linewidth]{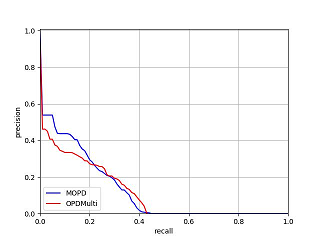}
  }
  \caption{\textbf{PR-curve}: Top left: mAP@IoU=0.5 for the detection of openable parts using bounding boxes. Top right: mAP@IoU=0.5 for the prediction of motion parameters using bounding boxes. Bottom left: mAP@IoU=0.5 for the detection of openable parts with masks. Bottom right: mAP@IoU=0.5 for the prediction of motion parameters with masks.}
  \label{fig:PR}
\end{figure*}

\subsection{Quantitative and Qualitative Results}

\begin{table}
\centering
\caption{ In the second row, the perceptual grouping encoder is replaced with SAM. In the third row, the geometric understanding encoder is replaced with Omnidata. In the third row. We frozen EfficientSAM and DSINE. }
\label{table:Alternative}
\resizebox{0.5\linewidth}{!}{%
\begin{tabular}{ccccc}
\toprule
Model         & PDet & +M & +MA & +MAO \\
\midrule
OPDMulti     & 32.9       & 31.6          & 19.4  & 16.0          \\
MOPD (with SAM)      & 34.3       & 33.1          & 19.6  & 15.9      \\
MOPD (with Omnidata)      & 33.5      &  32.3         &18.9   & 16.5 \\
MOPD (frozen) & 35.4       & 32.7          & 19.6  & 16.1                 \\
MOPD          & \bf 37.3       & \bf 36.1          & \bf 20.7  & \bf 16.6         \\
\bottomrule
\end{tabular}
}
\end{table}

As shown in Table \ref{table:Comparison of result}, our MOPD framework demonstrates notable superiority over the OPDMulti baselines across various metrics, improving part detection mAP by 4.9\% and motion parameter accuracy by 1.2\%, respectively. The remarkable performance of our approach can be attributed to several key factors, with one of the primary contributors being the integration of the perceptual grouping encoder and geometric understanding encoder. These encoders collaborate to extract essential features that significantly improve part detection accuracy. The perceptual grouping encoder identifies and segments different parts of the object, while the geometric understanding encoder captures geometric details and surface normal, providing additional context for enhanced detection. This integration improves part detection and enhances the accuracy of motion parameters. The extracted features enable a more precise understanding of the object's shape, see Fig.\ref{fig:ablation study SAM}. Resulting in more accurate motion parameter estimations see Fig. \ref{fig:ablation study normal}. 

\begin{table*}
\centering
\caption{Ablation study: It indicates that geometric features can not only enhance the prediction of motion parameters but also improve the ability of part detection. This is because pixels with the same surface normal are more likely to belong to a same part.}
\label{table:efficiency1}
\resizebox{\linewidth}{!}{%
\begin{tabular}{ccccccccc}
\toprule
Model         & PDet & +M & +MA & +MAO & Model Size & Test Memory & Training Memory & Computation time per image \\
\midrule
OPDMulti                  & 32.9     & 31.6     & 19.4  & 16.0  & 188MB  & 4942MB  & 175868 MB   &0.190s     \\
MOPD (w/o geometric)     & 34.3     & 33.1     & 19.6  & 15.9  & 224MB  & 5433MB  & 187393 MB   &0.207s     \\
MOPD (w/o perceptual)    & 36.7     & 35.4     & 20.2  & \textbf{16.7}  & 297MB  & 5768MB  & 190352 MB   &0.212s     \\
MOPD                      & \textbf{37.3}     & \textbf{36.1}     & \textbf{20.7}  & 16.6  & 371MB  & 5980MB  & 196308 MB   &0.228s     \\
\bottomrule
\end{tabular}
}
\end{table*}

\textbf{Qualitative results} are illustrated in Fig. \ref{fig:result}. It demonstrates the detection of a wide range of openable parts, a task also performed by OPDMulti. However, MOPD not only detects these parts but also accurately estimates their motion parameters. This capability is crucial in applications such as robotics and automation, where a precise understanding of object motion is vital for effective interaction and manipulation.

\textbf{PR-curves} of MOPD and OPDMulti are compared in Fig. \ref{fig:PR}. It consistently demonstrates that MOPD exhibits high precision at low recall rates. This implies that there are fewer false positives and false negatives at high thresholds. This indicates that our model performs better in detecting objects that more closely align with the definition of an openable part.

\textbf{Alternative pretrained models}
We first conducted experiments using SAM and Omnidata. Then, we found that EfficientSAM and DSINE are better. The result are shown in Tabel \ref{table:Alternative}.

\subsection{Optimal Transport}
\begin{figure}[htbp]
    \centering
     \begin{subfigure}[b]{.3\linewidth}
        \centering
        \includegraphics[scale=0.25]{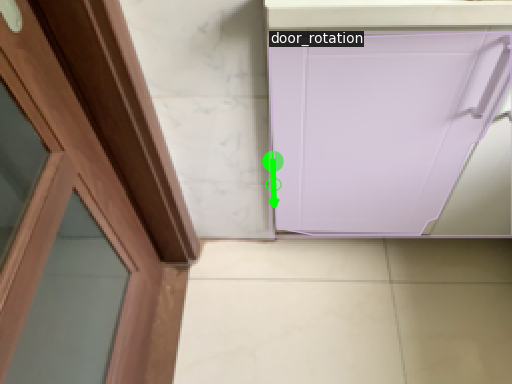} \\
        \includegraphics[scale=0.25]{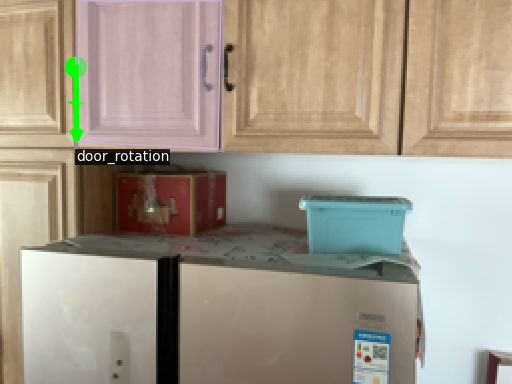} \\
        \includegraphics[scale=0.25]{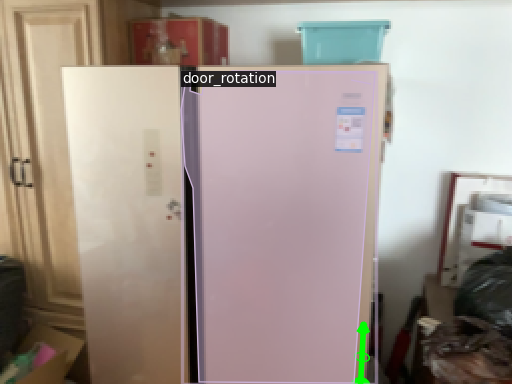} \\
        \includegraphics[scale=0.25]{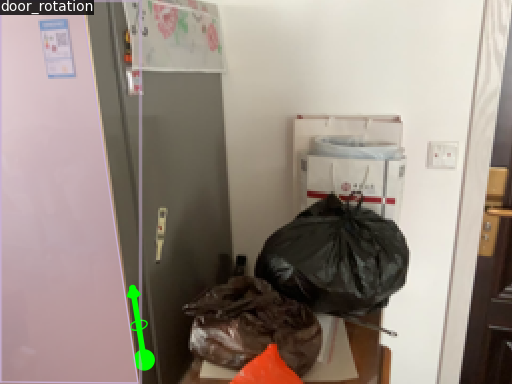}
        \caption{Ground Truth}
    \end{subfigure}
    \begin{subfigure}[b]{.3\linewidth}
        \centering
        \includegraphics[scale=0.25]{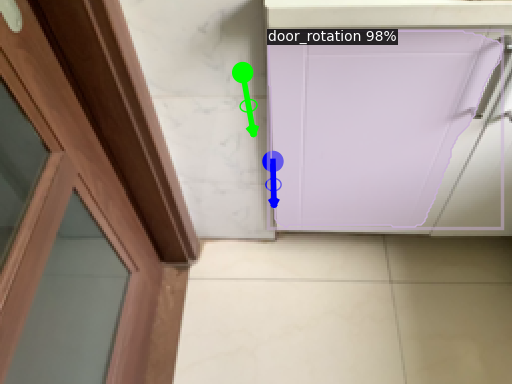} \\
        \includegraphics[scale=0.25]{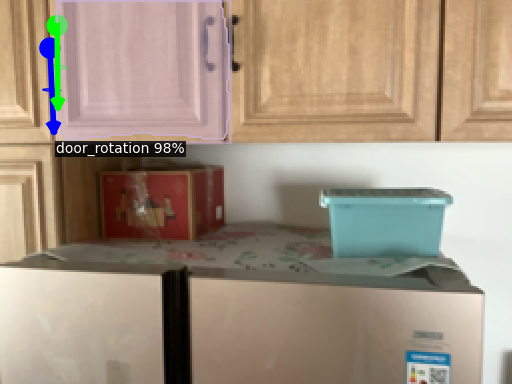} \\
        \includegraphics[scale=0.25]{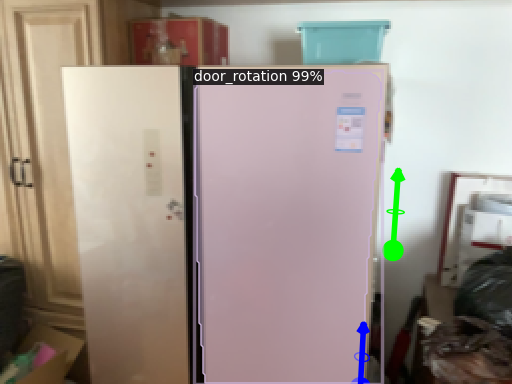} \\
        \includegraphics[scale=0.25]{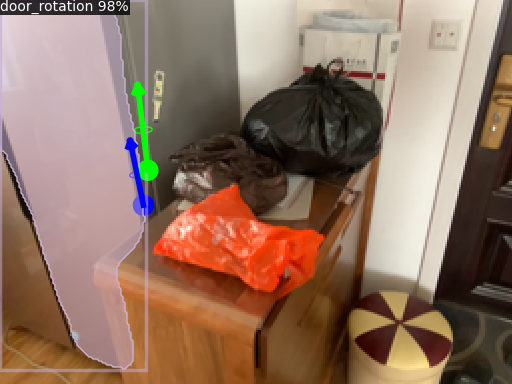} 
        \caption{MOPD(w/o OT)}
    \end{subfigure}
    \begin{subfigure}[b]{.3\linewidth}
        \centering
        \includegraphics[scale=0.25]{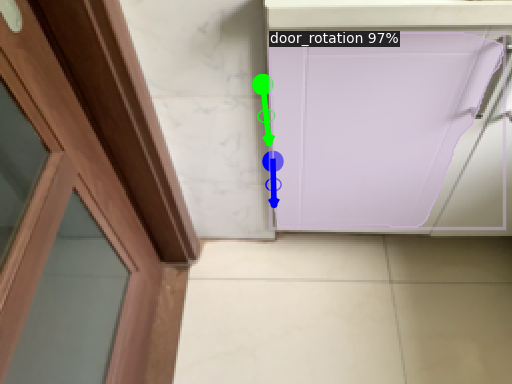} \\
        \includegraphics[scale=0.25]{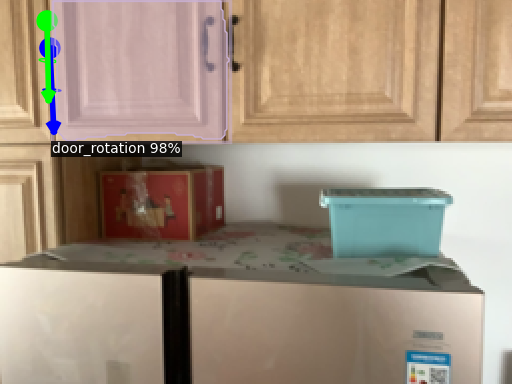} \\
        \includegraphics[scale=0.25]{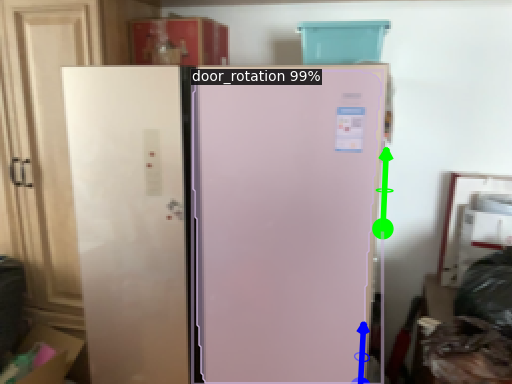} \\
        \includegraphics[scale=0.25]{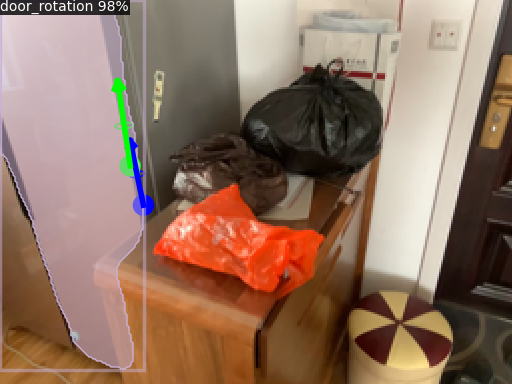}
        \caption{MOPD}
    \end{subfigure}
    \caption{\textbf{Qualitative Results of Ablation Study on Optimal Transport.} It is very intuitive that Optimal Transport's improvement in accurately perceiving the motion axis and origin for MOPD is very significant.}
    \label{fig:Qualitative Results of Ablation Study on Optimal Transport}
\end{figure}

As shown in Table. \ref{table:Comparison of result} and Figure. \ref{fig:Qualitative Results of Ablation Study on Optimal Transport}, our proposed motion match cost effectively improves the accuracy of motion axis and origin matching. The improvement brought by our method is due to ${C}_{origin}$ providing supervision for origin matching, thus limiting the drift of the motion origin. More importantly, ${C}_{axis}$ imposes constraints on the inclination of the motion axis, and more accurate motion trajectories lead to more accurate motion matching. 



\subsection{Efficiency}

The primary objective of openable part detection is to tackle the intricate challenge of enabling robotics to interact seamlessly with openable objects. Given the complexity involved in such interactions, efficiency becomes a paramount metric for evaluating the performance of any algorithm designed to achieve this goal. After all, an efficient algorithm can significantly enhance the robot's ability to perform tasks quickly and accurately.

To assess the efficiency of our approach, we conducted a series of experiments and compiled the results in Table \ref{table:efficiency1}. This table provides a comprehensive overview of how our model, equipped with DSINE and EfficientSAM, fares in terms of prediction accuracy and overall efficiency. As the table clearly illustrates, our model exhibits superior performance in both these aspects.

Notably, our approach also demonstrates that a smaller encoder can effectively address the OPD problem while maintaining a lightweight design. This is a significant advantage as it reduces the computational burden on the robot, enabling it to operate more efficiently and with less power consumption. In turn, this can lead to longer operational durations and fewer maintenance requirements, making our approach highly practical and appealing for real-world robotics applications.

\section{Conclusion}
\label{sec:conclusion}
Openable part detection plays a crucial role in applications involving interaction with articulated objects. This paper presents a two-stage OPD Transformer-based framework that integrates perceptual grouping and geometric features. In the initial stage, we introduce a perceptual grouping encoder to provide perceptual grouping feature priors for openable part detection, thereby improving detection outcomes through a cross-attention mechanism. Subsequently, in the second stage, a geometric understanding encoder offers geometric feature priors for detecting motion parameters. Finally, we introduced a motion cost in matching step, combined with the Optimal Transport model for training, which significantly improved the performance of the model. Extensive experiments demonstrate that our method outperforms previous approaches in both generalization and performance. Furthermore, ablation studies validate the effectiveness of the proposed models. We believe our method will bring benefits to downstream robotics applications.

\runinheading{Acknowledgments:}This work is supported  in part by the National Natural Science Foundation of China under Grant 62303405, in part by Ningbo Natural Science Foundation Project under Grant 2023J400, and in part by Ningbo Key Research and Development Plan under Grant 2023Z116.

\clearpage  

%
%
\bibliographystyle{splncs04}
\bibliography{main}
\end{document}